\documentclass{article}
\usepackage[preprint]{neurips_2026}
\makeatletter
\renewcommand{\@noticestring}{}
\makeatother
\usepackage[utf8]{inputenc}
\usepackage[T1]{fontenc}
\usepackage[hyphens]{url}
\usepackage{graphicx}
\usepackage{caption}
\usepackage{microtype}
\usepackage[hidelinks]{hyperref}
\usepackage{placeins}
\usepackage{float}
\usepackage{array}
\usepackage{booktabs}
\usepackage{amsmath}
\usepackage{amssymb}
\usepackage{algorithm}
\usepackage{algorithmic}
\newcommand{\ReFrame}{ReFrame}
\newcommand{\MSS}{MSS}
\newcommand{\LFF}{LFF}
\newcommand{\FBR}{FBR}

\title{\ReFrame{}: Silence-Aware Compute Reclamation for Full-Duplex Speech Models}
\author{
  Zheng Zhou\textsuperscript{1}\thanks{Equal contribution.},
  Guanyi Chen\textsuperscript{2,4,*},
  Jing Wei\textsuperscript{3,4},
  Haoyuan Liang\textsuperscript{1},
  Jian Zhang\textsuperscript{2,4},
  Haohuan Fu\textsuperscript{1}\\
  \textsuperscript{1}Tsinghua University\\
  \textsuperscript{2}The Hong Kong University of Science and Technology (Guangzhou)\\
  \textsuperscript{3}Sun Yat-sen University\\
  \textsuperscript{4}ScaleX-IO
}

\begin{document}
\raggedbottom
\maketitle

\begin{abstract}
Full-duplex speech models execute dense model inference for every
audio frame throughout a live interaction. This uniform framewise schedule
creates sustained GPU load across dialogue states whose computation has
strong, reusable temporal structure. \textbf{\ReFrame{}} formulates full-duplex
inference as a silence-aware computation problem. Mutual Silence Speculation
(\textsc{\MSS{}}) operates when both participants are silent: it
evaluates multi-frame Temporal continuations, verifies text and hidden-state
stability, and applies batched Mimi audio-token correction. Listening Frame
Folding (\textsc{\LFF{}}) operates while the user speaks and the agent listens:
it processes already-arrived user frames through parallel Temporal execution and
frame-batched audio-token generation while accepting only the verified prefix.
Frame-Budget Reinvestment
(\textsc{\FBR{}}) assigns the released frame intervals to resumable external
retrieval and language-model workers, then injects timely references
through Moshi's conditioning interface. We instantiate \ReFrame{} on five
full-duplex models spanning Moshi-family, phase-structured, and flattened
autoregressive generation. Across the complete 1,225-conversation FDBench,
\ReFrame{} reduces measured GPU execution time by 18.80--31.01\% and retains 96.92--100\% non-PAD text
agreement on five different models.
On HaluEval tests, \FBR{} increases the number of correct generated spoken
answers from 86 to 205 under the same auxiliary-service budget.
\ReFrame{} turns recurring
dialogue structure into lower inference cost and higher end-to-end service
capability.
\end{abstract}

\section{Introduction}
\begin{figure}[t]
\centering
\includegraphics[width=0.70\linewidth]{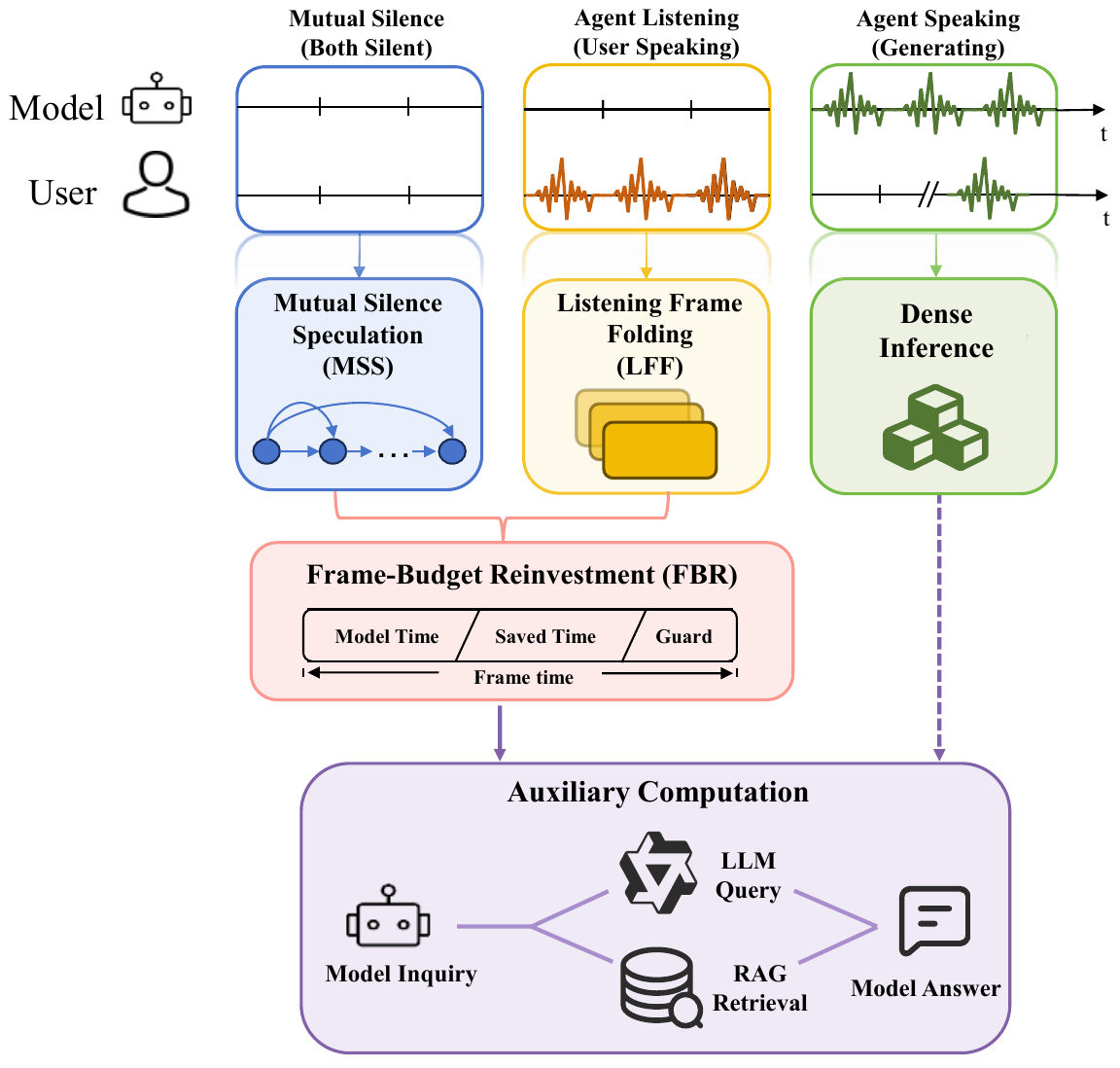}
\caption{\ReFrame{} routes mutual silence to \MSS{} and agent-listening
intervals to \LFF{}, while retaining dense execution during agent speech.
The frame budget released by \MSS{} and \LFF{} is reinvested through \FBR{}.}
\label{fig:reframe-overview}
\end{figure}
Spoken dialogue between users and intelligent agents is an important
human-computer interaction paradigm. Full-duplex models that simultaneously
generate responses and receive user input have attracted growing attention
because they support real-time barge-in and lightweight backchannels,
offering a conversational experience closer to human interaction
\citep{defossez2024moshispeechtextfoundationmodel,fang2026bayling,roy2026personaplexvoicerolecontrol}.

This capability, however, imposes strict per-frame latency constraints on
model inference. The system must produce one audio frame within a small time
budget. For example, Moshi emits one Mimi frame every 80 ms at 12.5 Hz. Meeting
this fixed frame rate with dense execution requires the complete inference
process at every frame and leaves little room for deeper reasoning or knowledge
access during the interaction.

A full-duplex model operating on speech as its primary modality also spends a
large fraction of every conversation in silence. In routine full-duplex
services, the agent remains silent for up to 84\% of interaction frames.
Existing systems have largely focused on using these silent intervals for
additional internal reasoning, such as chain-of-thought-style latent
computation, to improve response quality
\citep{wu2026silent,huang2026duplexomni}. These
approaches improve model capability through additional computation, but they
also increase inference cost and make per-frame latency constraints more
challenging to satisfy. The characteristic output regularity of silent
intervals provides a complementary opportunity to reduce foreground model
computation.

Moreover, prior work has mostly focused on the agent's ``listening'' phase, in
which the user speaks while the agent remains silent. The complementary regime
of mutual silence, where neither party speaks, has received far less attention.
We observe that these two silence regimes have different temporal structure and
therefore expose different opportunities for reducing inference cost.

\ReFrame{} converts both regimes into complementary, low-cost inference
paths. Figure~\ref{fig:reframe-overview} summarizes its state-aware execution
and compute-reinvestment workflow. For mutual silence, Mutual Silence
Speculation (\textsc{\MSS{}})
extends the recent silent state across a block of frames and verifies the
continuation with a multi-position Temporal pass. For the listening regime, Listening Frame
Folding (\textsc{\LFF{}}) buffers already-arrived user frames and evaluates
their Temporal positions and within-frame Depth Transformer steps together while preserving
autoregression across codebook levels. Both mechanisms reduce dense
per-frame model computation with almost no loss of output fidelity.

On the full FDBench set of 1,225 full-duplex spoken interactions,
\ReFrame{} reduces Joint GPU cost by 18.80--31.01\% across all five major full-duplex speech
models while retaining 99.33--100\% text agreement. Across the three Moshi
checkpoints, non-PAD text agreement is
96.92--98.12\%, while BayLing-Duplex
and PersonaPlex preserve exact trajectories across all 1,225 conversations.

Frame-Budget Reinvestment (\textsc{\FBR{}}) uses the compute reclaimed by
\textsc{\MSS{}} and \textsc{\LFF{}} for resumable auxiliary tasks and injects
timely references into ongoing full-duplex responses. Across five spoken
knowledge and reasoning evaluations, \FBR{} delivers more timely, correct
references. A HaluEval evaluation further shows an increase in
correct generated spoken answers from 86 to 205.

We introduce \textbf{\ReFrame{}} and make the following contributions.
\begin{itemize}
    \item We develop \MSS{}, which reuses verified state during mutual
    silence, and \LFF{}, which folds already-arrived user frames during the
    agent's listening state
    into a parallel multi-position block.
    \item We adapt and optimize these mechanisms across multiple
    full-duplex models, covering Moshi and Moshi-based multi-stream architectures
    with hierarchical audio-token generation and the multi-channel interleaved
    BayLing-Duplex sequence.
    \item We introduce \FBR{}, which
    converts the saved GPU budget into timely external retrieval and resident
    language-model computation, and injects the resulting references into
    ongoing full-duplex responses.
\end{itemize}

\section{Related Work}
\paragraph{Full-duplex speech models and silence structure.}
Moshi introduced a real-time speech-text architecture with hierarchical Mimi
audio-token generation, a Temporal Transformer, and an autoregressive Depth Transformer
(Depformer).
PersonaPlex builds
controllable conversational roles and voices on a related full-duplex
architecture.
BayLing-Duplex instead represents interleaved speech streams with a single
autoregressive language model. Synchronous LLMs
incorporate explicit time information into language-model generation to align
full-duplex responses with real-world interaction
\citep{veluri2024turnbasedinterfacessynchronousllms}, while OmniFlatten unifies speech and text in a
flattened sequence for low-latency full-duplex generation
\citep{zhang2025omniflattenendtoendgptmodel}. LSLM fuses listening and speaking
channels for real-time turn-taking
\citep{ma2024languagemodellistenspeaking},
SoulX-Duplug offloads dialogue management to a lightweight module
\citep{yan2026soulxduplugplugandplaystreamingstate}, and VoxMind extends
end-to-end spoken dialogue toward agentic interaction
\citep{liang2026voxmindendtoendagenticspoken}. Recent studies further examine
user-stream routing while speaking, synchronization, and turn-taking behavior
\citep{lu2026llmslistenspeakingstudy,riera2026synchronizationturntakingfullduplexspeech}.
Turn-taking research also establishes prosody and voice-activity projection as
useful signals for anticipating conversational transitions
\citep{ekstedt2022doesprosodyhelpturntaking,inoue2024multilingualturntakingpredictionusing}.
Full-Duplex-Bench and its overlap-focused extension provide complementary
evaluations of turn-taking and simultaneous speech
\citep{lin2025fullduplexbenchbenchmarkevaluatefullduplex,lin2026fullduplexbenchv15evaluatingoverlap}.
These models deliver highly interactive speech generation under a strict per-frame latency
budget, and their inputs arrive continuously so that inference remains
active at every audio frame. A large fraction of this interaction is spent in
silence, but existing systems treat mutual silence and listening identically.

\paragraph{Speculative and parallel decoding.}
Speculative decoding uses a smaller model to propose multiple candidate tokens
and the full model to verify them. Its key idea is to replace multiple
sequential decoding steps with one parallel verification pass, reducing the
target model overhead when several consecutive candidates are accepted
\citep{leviathan2023fastinferencetransformersspeculative,chen2023acceleratinglargelanguagemodel}.
Follow-up methods predict several future positions with additional decoding heads, as in
Medusa \citep{cai2024medusasimplellminference}, or use feature-level uncertainty
and adaptive candidate trees, as in EAGLE and EAGLE-2
\citep{li2025eaglespeculativesamplingrequires,li2024eagle2}. These methods target
autoregressive streams in which future output positions can be proposed and
verified against a fixed conditioning context. Extending this paradigm to
full-duplex speech introduces several constraints. Future user frames are
exogenous inputs and are unavailable before arrival. Speech generation also
couples semantic outputs with multiple dependent codec-token streams, which
makes exact agreement over a multi-frame proposal difficult to maintain. A
mismatch can affect both cross-frame Temporal states and within-frame
codec-token dependencies and may require state restoration and suffix replay
under a strict frame deadline. \ReFrame{} exploits the state-specific temporal regularities of full-duplex speech models.

\paragraph{Auxiliary computation in spoken dialogue.}
MoshiRAG studies asynchronous knowledge augmentation for full-duplex speech
models by invoking an auxiliary language model to improve factual responses
\citep{chien2026moshirag}. DuplexOmni adds listening, visual perception,
thinking, and speaking capabilities to real-time multimodal interaction. The
Silent Thought models latent internal reasoning for full-duplex spoken
dialogue to improve response quality. Retrieval-augmented language models
connect generation
to external corpora through learned retrieval, including REALM and RETRO
\citep{guu2020realmretrievalaugmentedlanguagemodel,borgeaud2022improvinglanguagemodelsretrieving}.
Dense Passage Retrieval, ColBERT, and unsupervised dense retrieval develop
effective candidate-search and ranking mechanisms
\citep{karpukhin2020densepassageretrievalopendomain,khattab2020colbertefficienteffectivepassage,gao2022precisezeroshotdenseretrieval},
while Fusion-in-Decoder integrates retrieved passages into generation
\citep{izacard2021leveragingpassageretrievalgenerative}. Adaptive-RAG, FLARE,
and IRCoT schedule retrieval according to question complexity, generation
uncertainty, or intermediate reasoning
\citep{jeong2024adaptiveraglearningadaptretrievalaugmented,jiang2023activeretrievalaugmentedgeneration,trivedi2023interleavingretrievalchainofthoughtreasoning}.
Systems such as FlexGen, Splitwise, and SGLang improve language-model serving
through memory-aware execution, phase specialization, and structured runtime
support
\citep{sheng2023flexgenhighthroughputgenerativeinference,patel2024splitwiseefficientgenerativellm,zheng2024sglangefficientexecutionstructured}.
These efforts focus on knowledge access, multimodal
capability, or response quality, and they typically increase rather than
reduce inference cost. They do not address how reclaimed foreground compute
can be safely reused within the per-frame latency budget.

\section{Methods}
\subsection{Preliminaries}
A full-duplex spoken dialogue model receives a user audio frame $x_t$ and
generates an agent output frame $y_t$ at each step of a fixed frame rate.
Its recurrent state $s_t$ carries the autoregressive caches and model context
used across frames. Dense inference evaluates the complete per-frame forward
and generation process:
\begin{equation}
(h_t,\ell_t,s_{t+1})=
\mathcal{T}_{\theta}(x_t,y_{t-1};s_t),
y_t=\mathcal{D}_{\theta}(h_t,\ell_t),
\end{equation}
where $h_t$ is the Temporal hidden state and $\ell_t$ denotes the text
prediction. In multi-stream architectures with hierarchical audio-token
generation, such as Moshi,
$\mathcal{T}_{\theta}$ is the Temporal Transformer that propagates cross-frame
context, while $\mathcal{D}_{\theta}$ is the within-frame autoregressive
Depth Transformer (Depformer) that generates Mimi audio tokens across codebook
levels. In multi-channel interleaved
architectures such as BayLing-Duplex, user speech, assistant text, and
assistant speech form one autoregressive sequence processed by a single LLM.

Let $u_t\in\{0,1\}$ and $a_t\in\{0,1\}$ denote user and agent speech activity.
Their activity pair defines the dialogue state used by the \ReFrame{} router.

\subsection{\ReFrame{} Overview}
The overall \ReFrame{} workflow routes each audio frame according to
the activity state of the user and the agent:
\begin{equation}
\pi(u_t,a_t)=
\begin{cases}
\textsc{\MSS{}}, & (u_t,a_t)=(0,0),\\
\textsc{\LFF{}}, & (u_t,a_t)=(1,0),\\
\textsc{Dense}, & \text{otherwise}.
\end{cases}
\end{equation}
Under mutual silence, \MSS{} proposes and verifies a block of silent
continuations, then batch-corrects tokens from their leading Mimi codebook
levels. During agent
listening, \LFF{} folds already-arrived user frames into masked multi-position
execution and releases verified outputs through a persistent output queue. \FBR{}
assigns the resulting GPU intervals to auxiliary computation within the frame
budget.

\subsection{Mutual Silence Speculation (\MSS{})}
During sustained mutual silence, when both the user and agent remain silent
for several frames, full-duplex models repeatedly emit a silent text token and
produce highly similar Temporal hidden states across neighboring
frames. Their Mimi audio tokens retain small frame-level variations, while
the decoded signal remains perceptually silent and preserves the same
interaction state. This combination yields a predictable silent-text
continuation, a relatively stable cross-frame representation, and benign
audio-token variation. \MSS{} turns this regularity into blockwise silent-state
speculation: it proposes several upcoming silent transitions together,
evaluates their Temporal positions in one multi-position block, and retains the
verified prefix.

\paragraph{Speculation and staged verification.}
For each mutual-silence interval, \MSS{} activates after both streams have
remained silent for several frames, the Temporal hidden states exhibit high
similarity, and user energy remains low. It repeats the latest verified silent
user input and silent text token to propose $K_R$ candidate frames,
including their text predictions, Temporal hidden states, and agent-side Mimi
audio tokens.
As mutual silence deepens, the $K_R$ window grows dynamically to capture
longer speculative spans and improve blockwise execution efficiency.

The $K_R$ candidate positions are evaluated in one Temporal block.
Verification scans the resulting text predictions and hidden states in temporal
order. A candidate joins the accepted prefix when its text remains in the
silent continuation and the cosine similarity between its hidden state and the verified silence
anchor exceeds the threshold. If position $j$ fails, positions
$1,\ldots,j-1$ form the accepted prefix and the remaining suffix returns to
dense framewise execution.

After Temporal acceptance, one batched Depformer pass evaluates all accepted
positions and replaces the proposed token values from the first $N_v$ codebook
levels with their computed values; for Moshi, $N_v=3$ corresponds to CB0--CB2.
These corrected token values become feedback for subsequent frame
computation. Depformer KV is local to codebook generation within one frame
and is reset across frames. Cross-frame history resides in the Temporal KV
cache. Correction retains the accepted speculative Temporal hidden states and
KV entries, without sequentially replaying them under the corrected tokens. The higher residual
codebook levels describe fine variations that remain perceptually silent in this
state, so they retain the proposed silence values. Staged verification
therefore combines exact silent text predictions, stable verified hidden
states, and batched partial audio-token correction.

\begin{algorithm}[t]
\caption{Mutual Silence Speculation (\MSS{})}
\label{alg:mss}
\small
\begin{algorithmic}[1]
\REQUIRE model state $S$, silence anchor $h_s$, proposal window $K_R$,
maximum window $K_R^{\max}$
\IF{$\textsc{StableMutualSilence}(S,h_s)$}
    \STATE $K_R \leftarrow
           \textsc{GrowWindow}(K_R,\textsc{SilenceDepth}(S),K_R^{\max})$
    \STATE $Q_{1:K_R} \leftarrow \textsc{SilentContinuation}(S,K_R)$
    \STATE $(L_{1:K_R},H_{1:K_R})
           \leftarrow \textsc{TemporalBlock}(S,Q_{1:K_R})$
    \STATE $a\leftarrow 0$
    \WHILE{$a<K_R$ and $L_{a+1}$ is silent and
           $\cos(H_{a+1},h_s)\geq\tau_h$}
        \STATE $a\leftarrow a+1$
    \ENDWHILE
    \STATE $C_{1:a} \leftarrow \textsc{DepformerBatch}(H_{1:a})$
    \STATE $F_{1:a} \leftarrow \textsc{Correct}(Q_{1:a},C_{1:a})$
    \STATE $\textsc{FrameQueue}(F_{1:a})$
    \RETURN $\textsc{ReleaseOnSilentArrivals}(F_{1:a})$
\ENDIF
\RETURN $\textsc{DenseStep}()$
\end{algorithmic}
\end{algorithm}

\paragraph{Framewise commitment and recovery.}
Each accepted frame waits in the candidate buffer. At every real user-frame arrival,
the router confirms that the observed input matches the silent continuation
and releases the corresponding cached agent frame. When user speech begins,
\MSS{} commits the already verified prefix, discards the remaining silent
candidates, and resumes dense framewise execution from the resulting state.

The same principle maps naturally to other generation structures. A
multi-channel interleaved model proposes a silent token block, verifies the
prefix exactly, and retains the verified positions. PersonaPlex uses its
stable period-two silence pattern to propose an alternating two-frame block,
then verifies the text token and all eight agent-side Mimi audio tokens at every
accepted position. Across architectures, \MSS{} converts a persistent
mutual-silence state into verified block execution.

\subsection{Listening Frame Folding (\LFF{})}
\begin{figure}[t]
\centering
\includegraphics[width=0.74\linewidth]{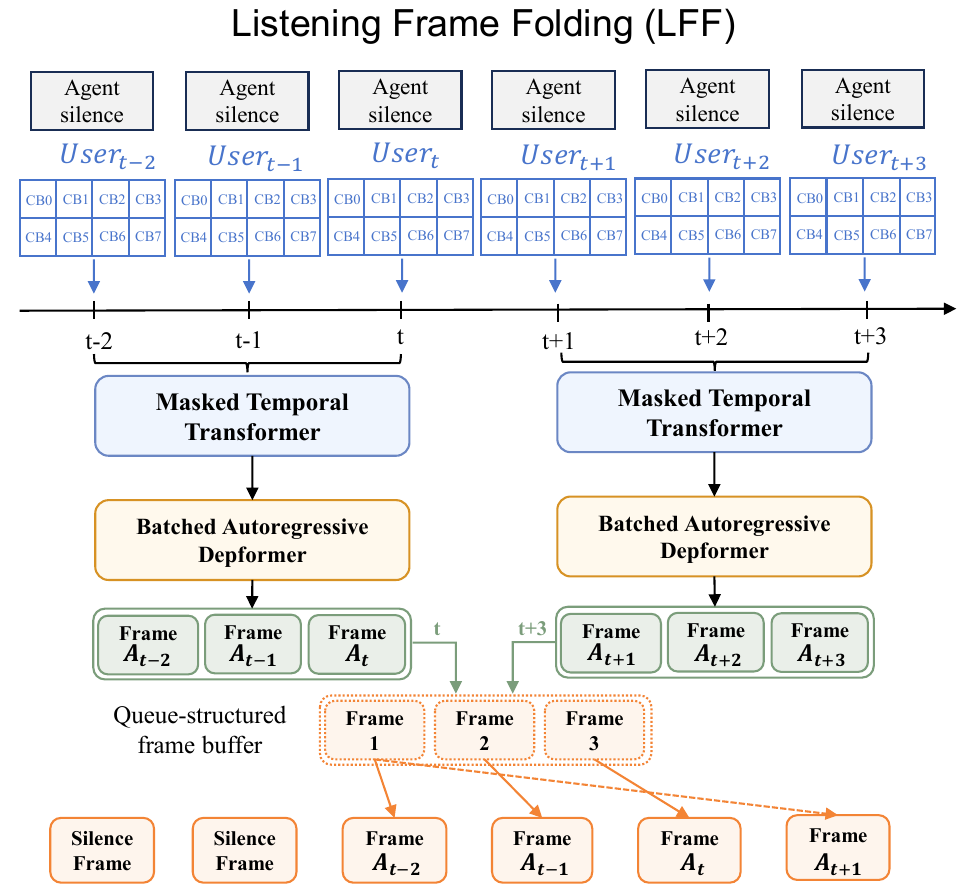}
\caption{\LFF{} folds already-arrived user-conditioned positions into masked
Temporal and frame-batched Depformer execution, then releases validated outputs
through a persistent output queue.}
\label{fig:lff-overview}
\end{figure}

When the user speaks while the agent listens, each frame contributes new user
audio. This continuously changing user input makes last-frame silence reuse
insufficient for prediction. The agent, however, commonly
maintains a silent listening state across several consecutive frames, creating
an opportunity to fold their computation. \LFF{} groups $K_L$ already-arrived
user frames, evaluates their cross-frame states in one masked multi-position
Temporal operation, generates candidate audio with one frame-batched Depformer
operation, and then validates the agent-side feedback. The
router enters \LFF{} after detecting sustained user speech together with agent
silence, and returns to framewise execution when the agent begins a non-silent
response or the listening interval ends. Figure~\ref{fig:lff-overview}
illustrates the folded execution and persistent output queue.

\paragraph{Persistent output queue.}
When \LFF{} first activates, it initializes an output queue that persists
throughout the session with $\delta=K_L-1$ copies of the latest verified silent
frame. These frames cover the output periods during which the first $K_L$ real
user Mimi frames arrive and form the first folded block. Because this
initialization briefly extends an agent segment that is already silent, its
perceptual effect is negligible. The completed block produces $K_L$ frames,
which are assembled according to the model's inter-codebook delay pattern and
appended to the queue together. The queue then releases them in chronological
order at the model's frame rate while the next real user frames form the
following folded block. Initialization occurs once per session; all subsequent
folded and framewise outputs enter the same persistent output queue.

\paragraph{Folded execution and validation.}
For a Moshi-style multi-stream model, let $x_i$ denote the observed user Mimi
frame at block position $i$, and let $q_i$ denote its provisional silent agent
feedback. \LFF{} evaluates all $K_L$ positions in one causally masked Temporal
operation:
\begin{equation}
\bigl(\hat h_i,\hat\ell_i\bigr)_{i=1}^{K_L}
=\mathcal{T}_{\theta}^{\mathrm{mask}}
\!\left(\{(x_i,q_{i-1})\}_{i=1}^{K_L};s_t\right).
\end{equation}
Every $x_i$ is already observed. Frame-batched Depformer execution generates
the candidate audio tokens before feedback validation. The verifier checks
the text token and all eight agent codebooks against the repeated feedback.
The causal mask preserves the dependency order of framewise execution. Let $r$ be the first
position whose computed agent feedback $y_r$ differs from $q_r$, or $K_L$ when
the complete block matches. Since position $i$ consumes the feedback from
position $i-1$, every position through $r$ receives the same user input and
preceding agent feedback as Dense. \LFF{} therefore accepts positions
$1{:}r$, restores the Temporal state at $r$, and replays only the suffix
$r+1{:}K_L$ from the corrected feedback.

The candidate audio tokens used in this check are computed as
\begin{equation}
z_{i,d}=\mathcal{D}_{\theta,d}
\!\left(\hat h_i,\hat\ell_i,z_{i,<d}\right),
\qquad i=1{:}K_L.
\end{equation}
Frame positions execute in parallel at each codebook depth $d$, while the
codebook depths remain autoregressive within each frame. A fully accepted
block consequently replaces $K_L$ Temporal calls and $K_L$ within-frame
generations with one multi-position Temporal operation and one frame-batched
Depformer generation. The verified and repaired frames supply the feedback
for subsequent model computation, while the persistent output queue controls
their ordered release.

For multi-channel interleaved models such as BayLing-Duplex, \LFF{} folds
packed per-frame token groups into one masked block and repairs the rejected
suffix in the unified autoregressive cache.

\subsection{Model-State Repair and Buffered Frame Output}
\ReFrame{} distinguishes the frame released as model output from the frame
stored as feedback for subsequent inference. Full-duplex models produce output
at a fixed frame rate; Moshi, for example, emits one frame every 80 ms at
12.5 Hz. A persistent output queue preserves this frame rate, while the
verified frame sequence supplies the text, audio tokens, and Temporal state used by
later computation.

\MSS{} caches its verified silent continuation and associates each cached
position with the corresponding subsequently arriving user frame. When that
user frame confirms continued silence, the cached agent frame becomes both the
released output and the frame committed to the autoregressive model state. At
user-speech onset, \MSS{}
commits the verified prefix, discards the remaining candidates, and restores
the Temporal cache at the prefix endpoint. \MSS{} therefore advances output
and autoregressive state with the same verified frame sequence.

\LFF{} uses this separation during its one-time queue initialization. The
$K_L-1$ added silent frames enter only the output queue and never become part
of the autoregressive context or Temporal feedback. Frames accepted by each folded Temporal
operation are committed as model feedback and appended to the queue; a
recomputed suffix follows the same path after repair. The initialization
therefore maintains silent output at the required frame rate without altering
the state that conditions subsequent dialogue.

\subsection{Frame-Budget Reinvestment (\FBR{})}
\MSS{} and \LFF{} shorten per-frame model execution while output frames continue
at the model's fixed frame rate. \FBR{} uses the GPU time remaining in each frame
period to advance an auxiliary task. Let $T_{\mathrm{frame}}$ denote the frame
period, $T_{\mathrm{model},t}$ the measured model execution time in frame $t$,
and $G_{w,t}$ a guard selected for worker $w$ according to its workload. The
GPU time available to the worker is
\begin{equation}
B_{w,t}=\max\!\left(0,T_{\mathrm{frame}}-T_{\mathrm{model},t}-G_{w,t}\right).
\end{equation}
Each worker divides a request into bounded, resumable chunks. Every frame runs
Moshi first; the scheduler then executes as many chunks as fit within
$B_{w,t}$ and carries the worker state into the next frame period. A
finalized request can replace unfinished older work at a chunk boundary.

\paragraph{RAG retrieval worker.}
During user speech, streaming transcription incrementally forms a query. The
worker begins candidate retrieval while the query is forming, refreshes the
candidates as additional words arrive, and finalizes the query when Moshi emits
\texttt{<ret>}. A first-stage search over a document corpus
produces candidates; a GPU reranker processes them in resumable blocks and
maintains the current top-$k$. The highest-ranked passages are compressed into
a reference and delivered through the MoshiRAG conditioning interface.

\paragraph{LLM-query worker.}
The LLM and Moshi remain resident on one GPU. Listening onset
starts prompt construction, and Moshi's \texttt{<ret>} token finalizes the
request. Prefill advances in token blocks, decoding advances token by token,
and the LLM KV cache preserves completed work across frame periods. The
completed result is delivered through the conditioning interface.

A completed reference is injected only if it arrives within the configured
conditioning window after the model issues the request. Unfinished or late
requests do not count as successful injections. Each chunk must fit within
the nonnegative budget before its guard deadline.

\section{Experiments}
\subsection{Experimental Setup}
\begin{table}[t]
\caption{Complete-FDBench efficiency and quality relative to Dense execution. Bold values mark the GPU cost reduction for each model.}
\label{tab:main}
\centering\small
\setlength{\tabcolsep}{3pt}
\textbf{(a) Execution cost and token agreement}\par\smallskip
\begin{tabular*}{\textwidth}{@{\extracolsep{\fill}}llrrrrr@{}}
\toprule
Model & Method & \shortstack{Time saved\\$\uparrow$} & Share & $n_c$ $\uparrow$ & Text $\uparrow$ & Exact $\uparrow$\\
\midrule
Moshiko & \MSS{} & 10.86\% & 46.49\% & 4.69 & 99.418\% & 527/1225\\
 & \LFF{} & 10.92\% & 40.27\% & 2.23 & 100\% & 1224/1225\\
 & Joint & \textbf{22.09\%} & 86.75\% & -- & 99.418\% & 527/1225\\
\midrule
Moshika & \MSS{} & 10.96\% & 46.52\% & 3.62 & 99.331\% & 505/1225\\
 & \LFF{} & 11.31\% & 40.14\% & 2.26 & 99.996\% & 1224/1225\\
 & Joint & \textbf{22.65\%} & 86.67\% & -- & 99.328\% & 504/1225\\
\midrule
Moshika-RL & \MSS{} & 7.52\% & 45.66\% & 3.39 & 99.446\% & 579/1225\\
 & \LFF{} & 10.92\% & 40.51\% & 2.23 & 100\% & 1225/1225\\
 & Joint & \textbf{18.80\%} & 86.19\% & -- & 99.446\% & 579/1225\\
\midrule
BayLing-Duplex & \MSS{} & 1.54\% & 11.12\% & 1.58 & 100\% & 1225/1225\\
 & \LFF{} & 28.62\% & 42.22\% & 4.20 & 100\% & 1225/1225\\
 & Joint & \textbf{31.01\%} & 53.34\% & -- & 100\% & 1225/1225\\
\midrule
PersonaPlex & \MSS{} & 17.42\% & 30.69\% & 3.25 & 100\% & 1225/1225\\
 & \LFF{} & 3.45\% & 36.80\% & 1.41 & 100\% & 1225/1225\\
 & Joint & \textbf{20.63\%} & 67.49\% & -- & 100\% & 1225/1225\\
\bottomrule
\end{tabular*}
\medskip
\textbf{(b) Speech fidelity and response onset}\par\smallskip
\begin{tabular*}{\textwidth}{@{\extracolsep{\fill}}llrrrrr@{}}
\toprule
Model & Method & CB $\Delta$ $\downarrow$ & WER$\Delta$ $\downarrow$ & Onset@3 $\uparrow$ & SNR $\uparrow$ & SI-SDR $\uparrow$\\
\midrule
Moshiko & \MSS{} & 0.66 & 3.94 & 98.13 & 92.73 & 92.64\\
 & \LFF{} & 0.00 & 0.00 & 100 & 104.24 & 104.24\\
 & Joint & 0.66 & 3.94 & 97.75 & 92.73 & 92.64\\
\midrule
Moshika & \MSS{} & 0.76 & 3.80 & 98.93 & 91.57 & 91.43\\
 & \LFF{} & 0.01 & 0.05 & 100 & 104.97 & 104.97\\
 & Joint & 0.76 & 3.84 & 98.28 & 91.49 & 91.34\\
\midrule
Moshika-RL & \MSS{} & 0.63 & 2.58 & 98.97 & 87.58 & 87.47\\
 & \LFF{} & 0.00 & 0.00 & 100 & 105.17 & 105.17\\
 & Joint & 0.63 & 2.58 & 98.31 & 87.58 & 87.47\\
\midrule
BayLing-Duplex & \MSS{} & 0 & -- & 100 & -- & --\\
 & \LFF{} & 0 & -- & 100 & -- & --\\
 & Joint & 0 & -- & 100 & -- & --\\
\midrule
PersonaPlex & \MSS{} & 0 & -- & 100 & -- & --\\
 & \LFF{} & 0 & -- & 100 & -- & --\\
 & Joint & 0 & -- & 100 & -- & --\\
\bottomrule
\end{tabular*}
\end{table}

\paragraph{Datasets.}
We evaluate frame-level inference cost and dialogue quality on the complete
FDBench set of 1,225 full-duplex spoken interactions
\citep{lin2025fullduplexbenchbenchmarkevaluatefullduplex,lin2026fullduplexbenchv15evaluatingoverlap}. FDBench covers nine
interaction patterns, including turn taking, interruption, backchannels,
overlap, and conversations in which the agent addresses another participant.
This diversity exposes both mutual-silence intervals and user-speaking
intervals under realistic changes in speaker activity. Our model suite
contains Moshiko and Moshika, Moshika-RL
\citep{ohashi2026multifacetedinteractivityalignment}, BayLing-Duplex, and
PersonaPlex, covering multiple full-duplex
architectures, generation policies, and model scales. We evaluate \FBR{} on
five spoken knowledge and reasoning workloads. The RAG-retrieval evaluation
uses HaluEval \citep{li2023haluevallargescalehallucinationevaluation},
OpenAudioBench \citep{li2025baichuanomni15technicalreport}, and WebQuestions
\citep{berant2013semanticparsingfreebase}, whereas the LLM-query evaluation
uses OpenAudioBench, SVAMP \citep{patel2021nlpmodelsreallyable}, and ASDiv
\citep{miao2021diversecorpusevaluatingdeveloping}. These
tasks span factual knowledge, audio-related facts, open-domain questions, and
arithmetic reasoning.

\paragraph{Comparison methods.}
Dense executes the complete forward and generation process at every
frame. \MSS{} enables mutual-silence speculation alone, \LFF{} enables listening
frame folding alone, and \textbf{Joint} routes between both mechanisms according to
the observed dialogue state. We compare all three methods against Dense in
both efficiency and quality. For \FBR{}, Dense and Joint run the same retrieval
worker on the same questions with the same waiting configuration and service
allocation. We compare actual reference injections and correct spoken answers.

\paragraph{Metrics.}
GPU cost reduction is the relative decrease in CUDA execution time measured
over the complete interaction. Share is the fraction of all interaction frames
in the state exploited by each mechanism: sustained mutual
silence for \MSS{}, user speech with a silent agent for \LFF{}, and their union
for Joint. The committed span $n_c$ is the mean number of frame positions that
each optimized multi-frame execution verifies and commits; it measures the
useful work obtained from one speculative or folded execution.
For BayLing-Duplex, each serialized 15-token output block represents five
logical frame positions, and $n_c$ is reported in these frame units.

Quality metrics use Dense as the reference. Text is the exact agreement of
agent text tokens at corresponding frame positions. Codebook
$\Delta$ is the fraction of all interaction frames in which Dense emits speech
and the optimized complete output-token frame differs from Dense. WER$\Delta$
is the normalized word edit distance between Whisper transcripts of the complete
Dense and optimized agent waveforms under the same transcription and text
normalization pipeline. Onset@3 is the fraction of Dense text-onset anchors whose
matched optimized onset is emitted within three frame periods. SNR and SI-SDR
are interaction-level dB averages over audio at exact non-PAD text-aligned
positions. The Moshi audio is decoded frame by frame with independent
streaming state per waveform, and WER uses Whisper-base with greedy decoding.
Exact reports the number
of interactions whose complete output-token trajectories match Dense after the
architecture-specific frame alignment. For \FBR{}, we count questions for which the model actually
consumes a reference and score the correctness of the resulting
spoken answers.

\paragraph{Settings.}
FDBench experiments run on NVIDIA A100-40GB GPUs. The \FBR{} experiment uses
Moshika-RAG, with the same
external reference-encoder service for Dense and Joint. Retrieval uses a
one-million-passage corpus containing Wikipedia passages and HaluEval
evidence. Qwen2.5-14B-Instruct \citep{qwen2025qwen25technicalreport}
scores transcribed spoken answers. GPU cost is measured using CUDA-event elapsed
time accumulated over the inference stages. Text,
codebook, and onset metrics compare paired Dense and optimized outputs.

\subsection{Full-FDBench Results}
\subsubsection{Overall Efficiency and Quality}
Table~\ref{tab:main} combines efficiency and quality over the complete FDBench benchmark.
Joint reduces measured GPU execution time by 22.09\%, 22.65\%, 18.80\%,
31.01\%, and 20.63\% on Moshiko, Moshika, Moshika-RL, BayLing-Duplex,
and PersonaPlex, respectively, establishing a consistent systems benefit across distinct full-duplex
architectures and generation policies. Its routed states cover 53.34--86.75\%
of interaction frames, so \ReFrame{} applies its optimized execution across a
large share of full-duplex interaction.

The contribution of each route follows the model's execution structure.
The three Moshi checkpoints save 7.52--10.96\% with \MSS{} and
10.92--11.31\% with \LFF{},
whereas BayLing-Duplex benefits primarily from folded listening blocks and
PersonaPlex from its regular mutual-silence pattern. These results demonstrate
that \ReFrame{} adapts its savings to architecture-specific dialogue structure.

Joint retains 99.328--100\% all-position text agreement. On the three Moshi
models, its full-frame differences during Dense speech affect 0.63--0.76\%
of all frames. WER differences are 2.58--3.94\%, matched-speech SNR is
87.58--92.73 dB, and SI-SDR is 87.47--92.64 dB.
BayLing-Duplex and PersonaPlex reproduce all 1,225 Dense
output-token trajectories exactly under alignment.
Joint's Onset@3 is 97.75\%, 98.28\%, and 98.31\% on the three Moshi models
and 100\% on BayLing-Duplex and PersonaPlex. \LFF{} achieves 100\%
Onset@3 on all five models.
\subsubsection{Ablation Study}
Table~\ref{tab:main} isolates the contribution of
each routed state. Across the evaluated models, \MSS{} and \LFF{} independently
reduce global GPU cost by up to 17.42\% and 28.62\%, respectively. Their verified-prefix
spans show that both mechanisms commit multiple useful frames per optimized
invocation on most architectures.

Joint improves global reduction on every model while retaining the quality of
both component paths, confirming that \MSS{} and \LFF{} contribute
independently and extend \ReFrame{} across complementary dialogue states.

\subsubsection{Simple Execution Baselines}
We compare three controls on Moshiko with the same backend and 1,225 inputs
(Table~\ref{tab:baselines}). Fixed-repeat speculation proposes six copies of
the silent frame and requires exact text and eight-codebook verification.
VAD bypass repeats the preceding output while skipping up to three forwards,
with a dense probe on the fourth frame and immediate dense execution on user
speech. Whole-block replay uses the same LFF queue and folded candidates,
but recomputes the entire block after a mismatch instead of retaining the
dependency-valid prefix.

\begin{table}[t]
\caption{Comparison with simple execution baselines on Moshiko.}
\label{tab:baselines}
\centering\small
\setlength{\tabcolsep}{8pt}
\begin{tabular}{@{}lrrr@{}}
\toprule
Execution policy & \shortstack[r]{Time\\saved} & \shortstack[r]{Non-PAD text\\agreement} & \shortstack[r]{CB\\difference}\\
\midrule
Fixed repeat, strict verification & 1.84 & 100.00 & 0.00\\
VAD bypass & 24.27 & 85.18 & 26.73\\
MSS & 10.86 & 96.92 & 5.35\\
Whole-block replay & 2.13 & 100.00 & 0.00\\
LFF & 10.92 & 100.00 & 0.00\\
Joint & 22.09 & 96.92 & 5.35\\
\bottomrule
\end{tabular}
\end{table}

Strict repeat verification preserves output fidelity but saves 1.84\% of
execution time. Joint achieves 22.09\% savings and substantially higher text
agreement than VAD bypass at a similar compute reduction. At equal queue
depth, LFF saves 10.92\% versus 2.13\% for whole-block replay. LFF handles
already-arrived, continuously changing user inputs during listening, covering
a dialogue state complementary to MSS. Its actual mean retained prefix is
2.23--2.26 frames across the three Moshi models, more than twice the one
valid position guaranteed by the dependency structure. Stable agent feedback
therefore exposes reuse even while the user's input changes.

\subsubsection{Hyperparameter Sensitivity}
\begin{table}[H]
\caption{Moshika \MSS{} hyperparameter comparison.}
\label{tab:mss-selection}
\centering
{\small
\setlength{\tabcolsep}{8pt}
\begin{tabular}{@{}rrrrrr@{}}
\toprule
\shortstack{$K_R$} & Dwell & GPU $\downarrow$ & $n_c$ $\uparrow$ & Text $\uparrow$ & A.$\Delta$ $\downarrow$\\
\midrule
4 & 16 & 10.40\% & 3.458 & 99.268\% & 0.819\%\\
\textbf{6} & \textbf{16}
 & \textbf{10.72\%} & \textbf{3.624} & \textbf{99.320\%} & \textbf{0.743\%}\\
8 & 16 & 10.97\% & 3.742 & 99.203\% & 0.835\%\\
6 & 12 & 11.39\% & 3.739 & 99.190\% & 0.874\%\\
6 & 20 & 9.47\% & 3.658 & 99.332\% & 0.748\%\\
\bottomrule
\end{tabular}
}
\end{table}

We compare multiple settings of the principal \MSS{} and \LFF{} parameters on
Moshika over the complete FDBench benchmark to characterize their effects on the efficiency
and quality of the corresponding optimized execution mode. For \MSS{}, dwell is
the number of consecutive mutual-silence frames required before speculation,
and $K_R$ is the initial proposal-window length. During execution, the window
doubles as mutual silence deepens until reaching a maximum of 16 frames.
Table~\ref{tab:mss-selection} shows that a smaller dwell triggers speculation
earlier and covers a broader but less stable portion of mutual silence: dwell
12 raises reduction to 11.39\%, but lowers text agreement and increases
codebook divergence. Conversely, setting dwell to 20 delays speculation until deeper, more stable silence
and improves text agreement, but reduces coverage and lowers the saving to 9.47\%.
A larger $K_R$ enables longer speculative blocks earlier and allows more
frames to be accepted per verification, as reflected by the increasing $n_c$, but distant
candidates are less stable; $K_R=8$ gives only
0.25 percentage points more reduction than $K_R=6$ with lower quality, while
$K_R=4$ sacrifices reduction without improving quality. The selected
setting with $K_R=6$ and dwell 16 balances
output quality and GPU cost reduction.

\begin{table}[H]
\caption{Moshika \LFF{} window comparison.}
\label{tab:lff-selection}
\centering
{\small
\setlength{\tabcolsep}{12pt}
\begin{tabular}{@{}rrrr@{}}
\toprule
$K_L$ & GPU $\downarrow$ & $n_c$ $\uparrow$ & Mean onset delay $\downarrow$\\
\midrule
2 & 7.25\% & 1.533 & 0.93 frames\\
\textbf{4} & \textbf{11.25\%} & \textbf{2.264} & \textbf{2.78 frames}\\
6 & 11.64\% & 2.755 & 4.62 frames\\
8 & 10.68\% & 3.088 & 6.45 frames\\
\bottomrule
\end{tabular}
}
\end{table}

Table~\ref{tab:lff-selection} compares $K_L$, the number of already-arrived listening
frames folded into each block. Increasing $K_L$ folds more frames per
invocation and increases the verified prefix committed by each execution, but
also increases the persistent output-queue depth and raises response-onset delay. Moving from
$K_L=4$ to 6 increases $n_c$ from 2.264 to 2.755 while adding only 0.39
percentage points of reduction and increasing mean onset delay from 2.78 to
4.62 frames; at $K_L=8$, longer repair and underfilled-block overheads reduce
the realized saving. The selected $K_L=4$ setting provides a strong
efficiency--response-timing trade-off.

\FloatBarrier
\subsection{Frame-Budget Reinvestment Evaluation}
\FBR{} uses the frame time reclaimed by \MSS{} and \LFF{} to advance
retrieval and LLM queries. Auxiliary workers resume computation in the
available frame intervals, retain unfinished progress across frames, and
supply timely references to ongoing spoken generation. We evaluate both
the auxiliary results delivered on time and the model's final spoken answers.
\begin{figure}[t]
\centering
\includegraphics[width=\columnwidth]{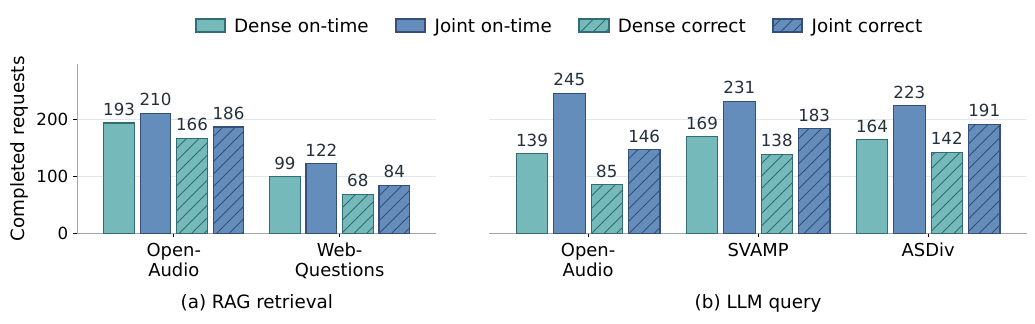}
\caption{\FBR{} results on two RAG retrieval and three LLM-query
workloads. Each dataset has four bars: on-time references for Dense and
Joint, followed by on-time-correct references for Dense and Joint. Each dataset has 300 questions.}
\label{fig:fbr-results}
\end{figure}

\paragraph{Timely auxiliary results.}
Figure~\ref{fig:fbr-results} shows that Joint converts reclaimed frame time
into more completed and correct references. Across the two RAG workloads,
Joint adds up to 23 on-time references and 20 on-time correct references;
across the three LLM-query workloads, it adds up to 106 on-time answers and 61
on-time correct answers. Every workload improves in both measures.

\paragraph{Final spoken answers.}
On 500 HaluEval questions, we track reference injection into Moshika-RAG
and evaluate the spoken answers generated by the model.
Table~\ref{tab:closed-loop} shows that Joint increases actual reference
injections from 112 to 356 and correct spoken answers from 86 to 205.
Spoken-answer accuracy rises from 17.20\% to 41.00\%, a gain of
23.80 percentage points.

\begin{table}[htbp]
\caption{Reference injections and generated spoken answers on 500 HaluEval questions.}
\label{tab:closed-loop}
\centering\small
\setlength{\tabcolsep}{4pt}
\begin{tabular}{lrr}
\toprule
Metric & Dense & Joint\\
\midrule
Questions with actual reference injection & 112 & 356\\
Correct generated spoken answers & 86 & 205\\
Spoken-answer accuracy (\%) & 17.20 & 41.00\\
\bottomrule
\end{tabular}
\end{table}

Together, these results show that \FBR{} turns lower frame-level inference
cost into both more timely external information and more correct final
spoken answers. Preserving auxiliary progress across frame intervals lets
retrieval and LLM workers use the reclaimed time to strengthen ongoing
full-duplex responses.

\FloatBarrier
\section{Conclusion}
\ReFrame{} establishes dialogue state as a direct source of systems
efficiency for full-duplex spoken interaction. Mutual Silence Speculation
(\MSS{}) converts persistent silence into verified Temporal blocks
with batched Mimi audio-token correction, while Listening Frame Folding (\LFF{})
converts already-arrived user frames during listening into parallel model execution.
Architecture-aware state verification and repair adapt these mechanisms to
Moshi and Moshi-based models with distinct silence patterns, as well as
multi-channel interleaved BayLing-Duplex. On the full
FDBench, their composition consistently reduces measured GPU cost across the
evaluated models while preserving near-identical text and high speech
fidelity. Frame-Budget Reinvestment (\FBR{}) turns the released frame
intervals into auxiliary computation, enabling the model to provide more
correct answers in a timely manner and thereby improve service quality.
These results connect lower frame-level inference cost
to higher end-to-end service capability, making state-aware computation a
practical foundation for efficient and knowledge-augmented full-duplex agents.

\bibliographystyle{abbrvnat}
\bibliography{references}

@misc{ohashi2026multifacetedinteractivityalignment,
      title={Multi-Faceted Interactivity Alignment in Full-Duplex Speech Models},
      author={Atsumoto Ohashi and Neil Zeghidour and Alexandre D{\'e}fossez and Eugene Kharitonov},
      year={2026},
      eprint={2606.11167},
      archivePrefix={arXiv},
      primaryClass={cs.CL},
      url={https://arxiv.org/abs/2606.11167},
}

@misc{defossez2024moshispeechtextfoundationmodel,
      title={Moshi: a speech-text foundation model for real-time dialogue},
      author={Alexandre Défossez and Laurent Mazaré and Manu Orsini and Amélie Royer and Patrick Pérez and Hervé Jégou and Edouard Grave and Neil Zeghidour},
      year={2024},
      eprint={2410.00037},
      archivePrefix={arXiv},
      primaryClass={eess.AS},
      url={https://arxiv.org/abs/2410.00037},
}

@misc{roy2026personaplexvoicerolecontrol,
      title={PersonaPlex: Voice and Role Control for Full Duplex Conversational Speech Models},
      author={Rajarshi Roy and Jonathan Raiman and Sang-gil Lee and Teodor-Dumitru Ene and Robert Kirby and Sungwon Kim and Jaehyeon Kim and Bryan Catanzaro},
      year={2026},
      eprint={2602.06053},
      archivePrefix={arXiv},
      primaryClass={cs.CL},
      url={https://arxiv.org/abs/2602.06053},
}

@misc{fang2026bayling,
      title={BayLing-Duplex: Native Full-Duplex Speech Dialogue with a Single Autoregressive LLM},
      author={Fang, Qingkai and Guo, Shoutao and Feng, Yang},
      year={2026},
      eprint={2606.14528},
      archivePrefix={arXiv},
      primaryClass={cs.CL},
      url={https://arxiv.org/abs/2606.14528},
}

@misc{leviathan2023fastinferencetransformersspeculative,
      title={Fast Inference from Transformers via Speculative Decoding},
      author={Yaniv Leviathan and Matan Kalman and Yossi Matias},
      year={2023},
      eprint={2211.17192},
      archivePrefix={arXiv},
      primaryClass={cs.LG},
      url={https://arxiv.org/abs/2211.17192},
}

@misc{chen2023acceleratinglargelanguagemodel,
      title={Accelerating Large Language Model Decoding with Speculative Sampling},
      author={Charlie Chen and Sebastian Borgeaud and Geoffrey Irving and Jean-Baptiste Lespiau and Laurent Sifre and John Jumper},
      year={2023},
      eprint={2302.01318},
      archivePrefix={arXiv},
      primaryClass={cs.CL},
      url={https://arxiv.org/abs/2302.01318},
}

@misc{cai2024medusasimplellminference,
      title={Medusa: Simple LLM Inference Acceleration Framework with Multiple Decoding Heads},
      author={Tianle Cai and Yuhong Li and Zhengyang Geng and Hongwu Peng and Jason D. Lee and Deming Chen and Tri Dao},
      year={2024},
      eprint={2401.10774},
      archivePrefix={arXiv},
      primaryClass={cs.LG},
      url={https://arxiv.org/abs/2401.10774},
}

@misc{li2025eaglespeculativesamplingrequires,
      title={EAGLE: Speculative Sampling Requires Rethinking Feature Uncertainty},
      author={Yuhui Li and Fangyun Wei and Chao Zhang and Hongyang Zhang},
      year={2025},
      eprint={2401.15077},
      archivePrefix={arXiv},
      primaryClass={cs.LG},
      url={https://arxiv.org/abs/2401.15077},
}

@misc{li2024eagle2,
      title={EAGLE-2: Faster Inference of Language Models with Dynamic Draft Trees},
      author={Li, Yuhui and Wei, Fangyun and Zhang, Chao},
      year={2024},
      eprint={2406.16858},
      archivePrefix={arXiv},
      primaryClass={cs.CL},
      url={https://arxiv.org/abs/2406.16858},
}

@misc{huang2026duplexomni,
      title={DuplexOmni: Real-Time Listening, Seeing, Thinking, and Speaking for Full-Duplex Interaction},
      author={Huang, Muye and Zhang, Lingling and Yu, Xingyu and Shi, Lei and Ma, Zhanyu and Xu, Jun and Gao, Jiuchong and Hao, Jinghua and He, Renqing and Liu, Jun},
      year={2026},
      eprint={2606.09186},
      archivePrefix={arXiv},
      primaryClass={cs.CL},
      url={https://arxiv.org/abs/2606.09186},
}

@misc{wu2026silent,
      title={The Silent Thought: Modeling Internal Cognition in Full-Duplex Spoken Dialogue Models via Latent Reasoning},
      author={Wu, Donghang and Zhang, Tianyu and Li, Yuxin and Liu, Hexin and Chen, Chen and Chng, Eng Siong and Bengio, Yoshua},
      year={2026},
      eprint={2603.17837},
      archivePrefix={arXiv},
      primaryClass={cs.CL},
      url={https://arxiv.org/abs/2603.17837},
}

@misc{chien2026moshirag,
      title={MoshiRAG: Asynchronous Knowledge Retrieval for Full-Duplex Speech Language Models},
      author={Chien, Chung-Ming and Orsini, Manu and Kharitonov, Eugene and Zeghidour, Neil and Livescu, Karen and D{\'e}fossez, Alexandre},
      year={2026},
      eprint={2604.12928},
      archivePrefix={arXiv},
      primaryClass={cs.CL},
      url={https://arxiv.org/abs/2604.12928},
}

@misc{li2023haluevallargescalehallucinationevaluation,
      title={HaluEval: A Large-Scale Hallucination Evaluation Benchmark for Large Language Models},
      author={Junyi Li and Xiaoxue Cheng and Wayne Xin Zhao and Jian-Yun Nie and Ji-Rong Wen},
      year={2023},
      eprint={2305.11747},
      archivePrefix={arXiv},
      primaryClass={cs.CL},
      url={https://arxiv.org/abs/2305.11747},
}

@misc{qwen2025qwen25technicalreport,
      title={Qwen2.5 Technical Report},
      author={Qwen and An Yang and Baosong Yang and Beichen Zhang and Binyuan Hui and Bo Zheng and Bowen Yu and Chengyuan Li and Dayiheng Liu and Fei Huang and Haoran Wei and Huan Lin and Jian Yang and Jianhong Tu and Jianwei Zhang and Jianxin Yang and Jiaxi Yang and Jingren Zhou and Junyang Lin and Kai Dang and Keming Lu and Keqin Bao and Kexin Yang and Le Yu and Mei Li and Mingfeng Xue and Pei Zhang and Qin Zhu and Rui Men and Runji Lin and Tianhao Li and Tianyi Tang and Tingyu Xia and Xingzhang Ren and Xuancheng Ren and Yang Fan and Yang Su and Yichang Zhang and Yu Wan and Yuqiong Liu and Zeyu Cui and Zhenru Zhang and Zihan Qiu},
      year={2025},
      eprint={2412.15115},
      archivePrefix={arXiv},
      primaryClass={cs.CL},
      url={https://arxiv.org/abs/2412.15115},
}

@misc{ma2024languagemodellistenspeaking,
      title={Language Model Can Listen While Speaking},
      author={Ziyang Ma and Yakun Song and Chenpeng Du and Jian Cong and Zhuo Chen and Yuping Wang and Yuxuan Wang and Xie Chen},
      year={2024},
      eprint={2408.02622},
      archivePrefix={arXiv},
      primaryClass={cs.CL},
      url={https://arxiv.org/abs/2408.02622},
}

@misc{yan2026soulxduplugplugandplaystreamingstate,
      title={SoulX-Duplug: Plug-and-Play Streaming State Prediction Module for Realtime Full-Duplex Speech Conversation},
      author={Ruiqi Yan and Wenxi Chen and Zhanxun Liu and Ziyang Ma and Haopeng Lin and Hanlin Wen and Hanke Xie and Jun Wu and Yuzhe Liang and Yuxiang Zhao and Pengchao Feng and Jiale Qian and Hao Meng and Yuhang Dai and Shunshun Yin and Ming Tao and Lei Xie and Kai Yu and Xinsheng Wang and Xie Chen},
      year={2026},
      eprint={2603.14877},
      archivePrefix={arXiv},
      primaryClass={eess.AS},
      url={https://arxiv.org/abs/2603.14877},
}

@misc{veluri2024turnbasedinterfacessynchronousllms,
      title={Beyond Turn-Based Interfaces: Synchronous LLMs as Full-Duplex Dialogue Agents},
      author={Bandhav Veluri and Benjamin N Peloquin and Bokai Yu and Hongyu Gong and Shyamnath Gollakota},
      year={2024},
      eprint={2409.15594},
      archivePrefix={arXiv},
      primaryClass={cs.CL},
      url={https://arxiv.org/abs/2409.15594},
}

@misc{zhang2025omniflattenendtoendgptmodel,
      title={OmniFlatten: An End-to-end GPT Model for Seamless Voice Conversation},
      author={Qinglin Zhang and Luyao Cheng and Chong Deng and Qian Chen and Wen Wang and Siqi Zheng and Jiaqing Liu and Hai Yu and Chaohong Tan and Zhihao Du and Shiliang Zhang},
      year={2025},
      eprint={2410.17799},
      archivePrefix={arXiv},
      primaryClass={cs.CL},
      url={https://arxiv.org/abs/2410.17799},
}

@misc{liang2026voxmindendtoendagenticspoken,
      title={VoxMind: An End-to-End Agentic Spoken Dialogue System},
      author={Tianle Liang and Yifu Chen and Shengpeng Ji and Yijun Chen and Zhiyang Jia and Jingyu Lu and Fan Zhuo and Xueyi Pu and Yangzhuo Li and Zhou Zhao},
      year={2026},
      eprint={2604.15710},
      archivePrefix={arXiv},
      primaryClass={cs.SD},
      url={https://arxiv.org/abs/2604.15710},
}

@misc{lu2026llmslistenspeakingstudy,
      title={How Should LLMs Listen While Speaking? A Study of User-Stream Routing in Full-Duplex Spoken Dialogue},
      author={Hui Lu and Xueyuan Chen and Huimeng Wang and Shuhai Peng and Shiyin Kang and Xixin Wu and Zhiyong Wu},
      year={2026},
      eprint={2605.10199},
      archivePrefix={arXiv},
      primaryClass={cs.CL},
      url={https://arxiv.org/abs/2605.10199},
}

@misc{riera2026synchronizationturntakingfullduplexspeech,
      title={Synchronization and Turn-Taking in Full-Duplex Speech Dialogue Models},
      author={Pablo Riera and Pablo Brusco and Cristina Kuo and Marcelo Sancinetti and S. R. K. Branavan},
      year={2026},
      eprint={2605.20356},
      archivePrefix={arXiv},
      primaryClass={cs.CL},
      url={https://arxiv.org/abs/2605.20356},
}

@misc{lin2025fullduplexbenchbenchmarkevaluatefullduplex,
      title={Full-Duplex-Bench: A Benchmark to Evaluate Full-duplex Spoken Dialogue Models on Turn-taking Capabilities},
      author={Guan-Ting Lin and Jiachen Lian and Tingle Li and Qirui Wang and Gopala Anumanchipalli and Alexander H. Liu and Hung-yi Lee},
      year={2025},
      eprint={2503.04721},
      archivePrefix={arXiv},
      primaryClass={cs.CL},
      url={https://arxiv.org/abs/2503.04721},
}

@misc{lin2026fullduplexbenchv15evaluatingoverlap,
      title={Full-Duplex-Bench v1.5: Evaluating Overlap Handling for Full-Duplex Speech Models},
      author={Guan-Ting Lin and Shih-Yun Shan Kuan and Qirui Wang and Jiachen Lian and Tingle Li and Shinji Watanabe and Hung-yi Lee},
      year={2026},
      eprint={2507.23159},
      archivePrefix={arXiv},
      primaryClass={eess.AS},
      url={https://arxiv.org/abs/2507.23159},
}

@misc{inoue2024multilingualturntakingpredictionusing,
      title={Multilingual Turn-taking Prediction Using Voice Activity Projection},
      author={Koji Inoue and Bing'er Jiang and Erik Ekstedt and Tatsuya Kawahara and Gabriel Skantze},
      year={2024},
      eprint={2403.06487},
      archivePrefix={arXiv},
      primaryClass={cs.CL},
      url={https://arxiv.org/abs/2403.06487},
}

@misc{guu2020realmretrievalaugmentedlanguagemodel,
      title={REALM: Retrieval-Augmented Language Model Pre-Training},
      author={Kelvin Guu and Kenton Lee and Zora Tung and Panupong Pasupat and Ming-Wei Chang},
      year={2020},
      eprint={2002.08909},
      archivePrefix={arXiv},
      primaryClass={cs.CL},
      url={https://arxiv.org/abs/2002.08909},
}

@misc{borgeaud2022improvinglanguagemodelsretrieving,
      title={Improving language models by retrieving from trillions of tokens},
      author={Sebastian Borgeaud and Arthur Mensch and Jordan Hoffmann and Trevor Cai and Eliza Rutherford and Katie Millican and George van den Driessche and Jean-Baptiste Lespiau and Bogdan Damoc and Aidan Clark and Diego de Las Casas and Aurelia Guy and Jacob Menick and Roman Ring and Tom Hennigan and Saffron Huang and Loren Maggiore and Chris Jones and Albin Cassirer and Andy Brock and Michela Paganini and Geoffrey Irving and Oriol Vinyals and Simon Osindero and Karen Simonyan and Jack W. Rae and Erich Elsen and Laurent Sifre},
      year={2022},
      eprint={2112.04426},
      archivePrefix={arXiv},
      primaryClass={cs.CL},
      url={https://arxiv.org/abs/2112.04426},
}

@misc{jeong2024adaptiveraglearningadaptretrievalaugmented,
      title={Adaptive-RAG: Learning to Adapt Retrieval-Augmented Large Language Models through Question Complexity},
      author={Soyeong Jeong and Jinheon Baek and Sukmin Cho and Sung Ju Hwang and Jong C. Park},
      year={2024},
      eprint={2403.14403},
      archivePrefix={arXiv},
      primaryClass={cs.CL},
      url={https://arxiv.org/abs/2403.14403},
}

@misc{ekstedt2022doesprosodyhelpturntaking,
      title={How Much Does Prosody Help Turn-taking? Investigations using Voice Activity Projection Models},
      author={Erik Ekstedt and Gabriel Skantze},
      year={2022},
      eprint={2209.05161},
      archivePrefix={arXiv},
      primaryClass={eess.AS},
      url={https://arxiv.org/abs/2209.05161},
}

@misc{jiang2023activeretrievalaugmentedgeneration,
      title={Active Retrieval Augmented Generation},
      author={Zhengbao Jiang and Frank F. Xu and Luyu Gao and Zhiqing Sun and Qian Liu and Jane Dwivedi-Yu and Yiming Yang and Jamie Callan and Graham Neubig},
      year={2023},
      eprint={2305.06983},
      archivePrefix={arXiv},
      primaryClass={cs.CL},
      url={https://arxiv.org/abs/2305.06983},
}

@misc{sheng2023flexgenhighthroughputgenerativeinference,
      title={FlexGen: High-Throughput Generative Inference of Large Language Models with a Single GPU},
      author={Ying Sheng and Lianmin Zheng and Binhang Yuan and Zhuohan Li and Max Ryabinin and Daniel Y. Fu and Zhiqiang Xie and Beidi Chen and Clark Barrett and Joseph E. Gonzalez and Percy Liang and Christopher Ré and Ion Stoica and Ce Zhang},
      year={2023},
      eprint={2303.06865},
      archivePrefix={arXiv},
      primaryClass={cs.LG},
      url={https://arxiv.org/abs/2303.06865},
}

@misc{patel2024splitwiseefficientgenerativellm,
      title={Splitwise: Efficient generative LLM inference using phase splitting},
      author={Pratyush Patel and Esha Choukse and Chaojie Zhang and Aashaka Shah and Íñigo Goiri and Saeed Maleki and Ricardo Bianchini},
      year={2024},
      eprint={2311.18677},
      archivePrefix={arXiv},
      primaryClass={cs.AR},
      url={https://arxiv.org/abs/2311.18677},
}

@misc{karpukhin2020densepassageretrievalopendomain,
      title={Dense Passage Retrieval for Open-Domain Question Answering},
      author={Vladimir Karpukhin and Barlas O{\u{g}}uz and Sewon Min and Patrick Lewis and Ledell Wu and Sergey Edunov and Danqi Chen and Wen-tau Yih},
      year={2020},
      eprint={2004.04906},
      archivePrefix={arXiv},
      primaryClass={cs.CL},
      url={https://arxiv.org/abs/2004.04906},
}

@misc{khattab2020colbertefficienteffectivepassage,
      title={ColBERT: Efficient and Effective Passage Search via Contextualized Late Interaction over BERT},
      author={Omar Khattab and Matei Zaharia},
      year={2020},
      eprint={2004.12832},
      archivePrefix={arXiv},
      primaryClass={cs.IR},
      url={https://arxiv.org/abs/2004.12832},
}

@misc{izacard2021leveragingpassageretrievalgenerative,
      title={Leveraging Passage Retrieval with Generative Models for Open Domain Question Answering},
      author={Gautier Izacard and Edouard Grave},
      year={2021},
      eprint={2007.01282},
      archivePrefix={arXiv},
      primaryClass={cs.CL},
      url={https://arxiv.org/abs/2007.01282},
}

@misc{gao2022precisezeroshotdenseretrieval,
      title={Precise Zero-Shot Dense Retrieval without Relevance Labels},
      author={Luyu Gao and Xueguang Ma and Jimmy Lin and Jamie Callan},
      year={2022},
      eprint={2212.10496},
      archivePrefix={arXiv},
      primaryClass={cs.IR},
      url={https://arxiv.org/abs/2212.10496},
}

@misc{trivedi2023interleavingretrievalchainofthoughtreasoning,
      title={Interleaving Retrieval with Chain-of-Thought Reasoning for Knowledge-Intensive Multi-Step Questions},
      author={Harsh Trivedi and Niranjan Balasubramanian and Tushar Khot and Ashish Sabharwal},
      year={2023},
      eprint={2212.10509},
      archivePrefix={arXiv},
      primaryClass={cs.CL},
      url={https://arxiv.org/abs/2212.10509},
}

@misc{zheng2024sglangefficientexecutionstructured,
      title={SGLang: Efficient Execution of Structured Language Model Programs},
      author={Lianmin Zheng and Liangsheng Yin and Zhiqiang Xie and Chuyue Sun and Jeff Huang and Cody Hao Yu and Shiyi Cao and Christos Kozyrakis and Ion Stoica and Joseph E. Gonzalez and Clark Barrett and Ying Sheng},
      year={2024},
      eprint={2312.07104},
      archivePrefix={arXiv},
      primaryClass={cs.AI},
      url={https://arxiv.org/abs/2312.07104},
}

@misc{li2025baichuanomni15technicalreport,
      title={Baichuan-Omni-1.5 Technical Report},
      author={Yadong Li and Jun Liu and Tao Zhang and Tao Zhang and Song Chen and Tianpeng Li and Zehuan Li and Lijun Liu and Lingfeng Ming and Guosheng Dong and Da Pan and Chong Li and Yuanbo Fang and Dongdong Kuang and Mingrui Wang and Chenglin Zhu and Youwei Zhang and Hongyu Guo and Fengyu Zhang and Yuran Wang and Bowen Ding and Wei Song and Xu Li and Yuqi Huo and Zheng Liang and Shusen Zhang and Xin Wu and Shuai Zhao and Linchu Xiong and Yozhen Wu and Jiahui Ye and Wenhao Lu and Bowen Li and Yan Zhang and Yaqi Zhou and Xin Chen and Lei Su and Hongda Zhang and Fuzhong Chen and Xuezhen Dong and Na Nie and Zhiying Wu and Bin Xiao and Ting Li and Shunya Dang and Ping Zhang and Yijia Sun and Jincheng Wu and Jinjie Yang and Xionghai Lin and Zhi Ma and Kegeng Wu and Jia li and Aiyuan Yang and Hui Liu and Jianqiang Zhang and Xiaoxi Chen and Guangwei Ai and Wentao Zhang and Yicong Chen and Xiaoqin Huang and Kun Li and Wenjing Luo and Yifei Duan and Lingling Zhu and Ran Xiao and Zhe Su and Jiani Pu and Dian Wang and Xu Jia and Tianyu Zhang and Mengyu Ai and Mang Wang and Yujing Qiao and Lei Zhang and Yanjun Shen and Fan Yang and Miao Zhen and Yijie Zhou and Mingyang Chen and Fei Li and Chenzheng Zhu and Keer Lu and Yaqi Zhao and Hao Liang and Youquan Li and Yanzhao Qin and Linzhuang Sun and Jianhua Xu and Haoze Sun and Mingan Lin and Zenan Zhou and Weipeng Chen},
      year={2025},
      eprint={2501.15368},
      archivePrefix={arXiv},
      primaryClass={cs.CL},
      url={https://arxiv.org/abs/2501.15368},
}

@misc{patel2021nlpmodelsreallyable,
      title={Are NLP Models really able to Solve Simple Math Word Problems?},
      author={Arkil Patel and Satwik Bhattamishra and Navin Goyal},
      year={2021},
      eprint={2103.07191},
      archivePrefix={arXiv},
      primaryClass={cs.CL},
      url={https://arxiv.org/abs/2103.07191},
}

@misc{miao2021diversecorpusevaluatingdeveloping,
      title={A Diverse Corpus for Evaluating and Developing English Math Word Problem Solvers},
      author={Shen-Yun Miao and Chao-Chun Liang and Keh-Yih Su},
      year={2021},
      eprint={2106.15772},
      archivePrefix={arXiv},
      primaryClass={cs.AI},
      url={https://arxiv.org/abs/2106.15772},
}

@misc{berant2013semanticparsingfreebase,
      title={Semantic Parsing on Freebase from Question-Answer Pairs},
      author={Jonathan Berant and Andrew Chou and Roy Frostig and Percy Liang},
      year={2013},
      howpublished={Proceedings of the 2013 Conference on Empirical Methods in Natural Language Processing},
      pages={1533--1544},
      doi={10.18653/v1/D13-1160},
      url={https://aclanthology.org/D13-1160/},
}

\clearpage
\appendix
\section*{Appendix}
\section{FDBench Composition and Metrics}
FDBench contains nine categories that exercise complementary full-duplex
interaction patterns. Table~\ref{tab:fdbench-tasks} gives their full names and
conversation counts. All results use the complete 1,225-conversation benchmark
and replay the same interaction traces for Dense and Joint.

\begin{table}[H]
\caption{FDBench categories and conversation counts.}
\label{tab:fdbench-tasks}
\centering
\small
\setlength{\tabcolsep}{6pt}
\begin{tabular}{@{}llr@{}}
\toprule
Abbr. & FDBench category & Conversations\\
\midrule
CPH & Candor Pause Handling & 216\\
CTT & Candor Turn Taking & 119\\
IBC & ICC Backchannel & 55\\
SPH & Synthetic Pause Handling & 137\\
SUI & Synthetic User Interruption & 200\\
BGS & Background Speech & 100\\
TTO & Talking to Other & 100\\
UBC & User Backchannel & 98\\
UI & User Interruption & 200\\
\midrule
All & Complete FDBench & 1,225\\
\bottomrule
\end{tabular}
\end{table}

The per-category metrics follow the definitions in the main paper. GPU-cost
reduction is the relative reduction in summed CUDA execution time over all
interactions in a category. Text is the exact agreement of agent text tokens
at corresponding frame positions. Codebook $\Delta$ is the percentage of all
interaction frames for which Dense is producing speech and the complete output
token frame differs from Dense. WER$\Delta$ is the normalized word-edit
distance between Whisper transcripts of Dense and Joint after applying the
same Whisper-base transcription and text-normalization procedure. Onset@3 is
the fraction of Dense text-onset anchors reproduced within three frame periods.
SNR is the interaction-level dB average over exact non-PAD text-aligned audio,
decoded frame by frame with independent streaming state per waveform. Exact counts an
interaction when its complete output-token trajectory matches Dense after the
architecture-specific frame alignment.

\subsection{Aggregation Details}
For a category $c$, let $\mathcal{I}_c$ be its interactions and let
$T_i^{\mathrm{D}}$ and $T_i^{\mathrm{J}}$ denote the accumulated Dense and
Joint CUDA execution time for interaction $i$. The reported reduction is
computed after summing time across the category:
\begin{equation}
R_c=1-\frac{\sum_{i\in\mathcal{I}_c}T_i^{\mathrm{J}}}
{\sum_{i\in\mathcal{I}_c}T_i^{\mathrm{D}}}.
\end{equation}
Text similarly divides the number of matching agent text-token positions by
the number of corresponding positions in the category. Codebook $\Delta$
uses the number of changed complete token frames during Dense speech as its
numerator and all interaction frames as its denominator. WER$\Delta$ sums
Whisper substitutions, deletions, and insertions before dividing by the total
number of words in the Dense transcripts. Onset@3 divides the number of Dense
onsets with a temporally corresponding Joint onset within three frames by the
total number of Dense onsets. These count-based aggregations reproduce the
same denominators used for the complete-FDBench values in the main paper.
For SNR, each interaction contributes one dB value, with signal and waveform-error
energies floored at $10^{-8}$ before taking their ratio. We average over
interactions with matched speech.

Moshi-family outputs are assembled according to their delayed Mimi-codebook
pattern before frame and speech-waveform comparison. BayLing-Duplex groups
each serialized 15-token block into five logical frames. PersonaPlex aligns
its period-two output phase before exact comparison. These conversions change
only the representation used by the evaluator; all models retain their native
generation order during execution.

\FloatBarrier
\section{Joint Results by FDBench Category}

Table~\ref{tab:task-cost} reports Joint GPU-cost
reduction for every model and category. Savings appear across all nine interaction patterns, including
pause handling, backchannels, interruptions, background speech, and speech
addressed to another participant. Tables~\ref{tab:moshiko-task-quality}--
\ref{tab:moshika-rl-task-quality} give the corresponding quality measurements
for the three Moshi checkpoints. Table~\ref{tab:exact-task-quality} reports the
per-category exact trajectories for BayLing-Duplex and PersonaPlex.

\begin{table}[htbp]
\caption{Joint GPU-cost reduction (\%) across all nine FDBench categories.}
\label{tab:task-cost}
\centering
\small
\setlength{\tabcolsep}{2.5pt}
\begin{tabular*}{\textwidth}{@{\extracolsep{\fill}}lrrrrrrrrr@{}}
\toprule
Model & CPH & CTT & IBC & SPH & SUI & BGS & TTO & UBC & UI\\
\midrule
Moshiko & 26.37 & 24.33 & 26.19 & 17.33 & 14.50 & 26.80 & 22.79 & 32.07 & 20.09\\
Moshika & 27.57 & 27.12 & 22.95 & 19.01 & 14.02 & 27.23 & 23.17 & 30.06 & 22.01\\
Moshika-RL & 25.33 & 24.01 & 25.42 & 20.41 & 13.40 & 16.01 & 14.94 & 18.99 & 16.99\\
BayLing-Duplex & 58.35 & 44.98 & 63.94 & 62.16 & 13.19 & 7.18 & 7.16 & 9.28 & 14.19\\
PersonaPlex & 10.50 & 20.42 & 4.42 & 11.91 & 18.62 & 36.14 & 32.73 & 45.96 & 23.09\\
\bottomrule
\end{tabular*}
\end{table}

\begin{table}[htbp]
\caption{Moshiko Joint quality by category. SNR is in dB; other values are percentages.}
\label{tab:moshiko-task-quality}
\centering
\small
\setlength{\tabcolsep}{6pt}
\begin{tabular}{@{}lrrrrr@{}}
\toprule
Task & Text $\uparrow$ & CB $\Delta$ $\downarrow$ & WER$\Delta$ $\downarrow$ & O@3 $\uparrow$ & SNR $\uparrow$\\
\midrule
CPH & 100.00 & 0.00 & 0.00 & 100.00 & 103.40\\
CTT & 99.84 & 0.24 & 1.48 & 99.33 & 99.60\\
IBC & 100.00 & 0.00 & 0.00 & 100.00 & 104.41\\
SPH & 100.00 & 0.00 & 0.00 & 100.00 & 103.91\\
SUI & 99.20 & 0.96 & 5.22 & 97.33 & 84.64\\
BGS & 99.57 & 0.40 & 3.31 & 97.91 & 91.98\\
TTO & 99.16 & 0.78 & 4.91 & 97.37 & 74.63\\
UBC & 99.90 & 0.11 & 1.31 & 99.34 & 101.04\\
UI & 98.43 & 1.81 & 7.67 & 95.11 & 79.68\\
\bottomrule
\end{tabular}
\end{table}

\begin{table}[htbp]
\caption{Moshika Joint quality by category. SNR is in dB; other values are percentages.}
\label{tab:moshika-task-quality}
\centering
\small
\setlength{\tabcolsep}{6pt}
\begin{tabular}{@{}lrrrrr@{}}
\toprule
Task & Text $\uparrow$ & CB $\Delta$ $\downarrow$ & WER$\Delta$ $\downarrow$ & O@3 $\uparrow$ & SNR $\uparrow$\\
\midrule
CPH & 100.00 & 0.00 & 0.00 & 100.00 & 104.16\\
CTT & 99.80 & 0.33 & 1.93 & 99.35 & 95.74\\
IBC & 100.00 & 0.00 & 0.00 & 100.00 & 104.86\\
SPH & 100.00 & 0.00 & 0.00 & 100.00 & 104.62\\
SUI & 98.77 & 1.32 & 5.86 & 97.06 & 77.81\\
BGS & 99.64 & 0.40 & 4.10 & 99.13 & 93.98\\
TTO & 98.82 & 1.45 & 6.32 & 96.95 & 72.22\\
UBC & 100.00 & 0.00 & 0.00 & 100.00 & 103.51\\
UI & 98.52 & 1.68 & 6.17 & 97.17 & 78.79\\
\bottomrule
\end{tabular}
\end{table}

\begin{table}[htbp]
\caption{Moshika-RL Joint quality by category. SNR is in dB; other values are percentages.}
\label{tab:moshika-rl-task-quality}
\centering
\small
\setlength{\tabcolsep}{6pt}
\begin{tabular}{@{}lrrrrr@{}}
\toprule
Task & Text $\uparrow$ & CB $\Delta$ $\downarrow$ & WER$\Delta$ $\downarrow$ & O@3 $\uparrow$ & SNR $\uparrow$\\
\midrule
CPH & 100.00 & 0.00 & 0.00 & 100.00 & 104.24\\
CTT & 99.95 & 0.06 & 0.12 & 99.79 & 101.31\\
IBC & 100.00 & 0.00 & 0.00 & 100.00 & 105.71\\
SPH & 100.00 & 0.00 & 0.00 & 100.00 & 104.37\\
SUI & 98.97 & 1.21 & 3.69 & 97.69 & 67.57\\
BGS & 99.60 & 0.48 & 2.25 & 98.28 & 78.77\\
TTO & 99.22 & 0.93 & 3.44 & 95.97 & 71.95\\
UBC & 99.75 & 0.25 & 2.83 & 99.13 & 97.29\\
UI & 98.80 & 1.29 & 4.28 & 97.48 & 72.43\\
\bottomrule
\end{tabular}
\end{table}

\begin{table}[htbp]
\caption{Joint exact output trajectories by FDBench category.}
\label{tab:exact-task-quality}
\centering
\small
\setlength{\tabcolsep}{6pt}
\begin{tabular}{@{}lcc@{}}
\toprule
Task & BayLing-Duplex & PersonaPlex\\
\midrule
CPH & 216/216 & 216/216\\
CTT & 119/119 & 119/119\\
IBC & 55/55 & 55/55\\
SPH & 137/137 & 137/137\\
SUI & 200/200 & 200/200\\
BGS & 100/100 & 100/100\\
TTO & 100/100 & 100/100\\
UBC & 98/98 & 98/98\\
UI & 200/200 & 200/200\\
\bottomrule
\end{tabular}
\end{table}

The three Moshi checkpoints retain high text agreement across every category,
with the largest token differences concentrated in interruption-heavy
interactions. Their Onset@3 values remain high across the same categories.
BayLing-Duplex and PersonaPlex reproduce all 1,225 Dense trajectories exactly,
including every conversation in each individual category.

\FloatBarrier
\section{Main Execution Parameters}
Tables~\ref{tab:mss-hparams} and~\ref{tab:lff-hparams} give the principal
parameters used for the complete-FDBench runs in Table~\ref{tab:main}.
For \MSS{}, \emph{dwell} is the number of consecutive mutual-silence frames
required before speculation. The hidden threshold is the minimum cosine
similarity between a candidate Temporal state and the verified silence anchor.
The return-hazard threshold limits speculation to intervals whose estimated
near-term return-to-speech risk remains low. For the Moshi checkpoints,
$K_R$ is the initial proposal span and grows as mutual silence deepens. The
codebook entry specifies how many leading Mimi codebook levels are recomputed
in the batched Depformer correction after text-and-hidden acceptance.

\begin{table}[htbp]
\caption{Principal \MSS{} parameters for the complete-FDBench runs.}
\label{tab:mss-hparams}
\centering
\small
\setlength{\tabcolsep}{3pt}
\begin{tabular}{@{}lp{0.67\columnwidth}@{}}
\toprule
Model & \MSS{} configuration\\
\midrule
Moshiko & Dwell 16; hidden similarity 0.990; return hazard 0.050; $K_R=6$; batched CB0--CB2 correction.\\
Moshika & Dwell 16; hidden similarity 0.995; return hazard 0.229; $K_R=6$; batched CB0--CB2 correction.\\
Moshika-RL & Dwell 24; hidden similarity 0.995; r95 return-hazard calibration with threshold 0.465; $K_R=6$; batched CB0--CB2 correction.\\
BayLing-Duplex & User-silence threshold 0.005; 15-token repeat chunk representing five logical frames; exact-prefix verification.\\
PersonaPlex & Period-two silence block; dwell 4; eight-frame proposal window; text and all eight agent codebooks verified.\\
\bottomrule
\end{tabular}
\end{table}

For \LFF{}, $K_L$ is the number of already-arrived user-conditioned frame
positions folded into one masked Temporal operation. A Moshi output queue is
initialized once with $K_L-1$ silent frames, so $K_L=4$ gives a queue depth of
three frames. Lazy activation creates this queue at the first eligible
listening interval. Exact-prefix commit retains the causally valid prefix and
replays only the suffix after the first feedback mismatch. The PersonaPlex
codebook margins are numerical equality tolerances used by its FP32
architecture-specific verifier.

\begin{table}[htbp]
\caption{Principal \LFF{} parameters for the complete-FDBench runs.}
\label{tab:lff-hparams}
\centering
\small
\setlength{\tabcolsep}{3pt}
\begin{tabular}{@{}lp{0.67\columnwidth}@{}}
\toprule
Model & \LFF{} configuration\\
\midrule
Moshiko & $K_L=4$; queue depth 3; persistent frame queue; prefix validation and suffix repair.\\
Moshika & $K_L=4$; queue depth 3; persistent frame queue; prefix validation and suffix repair.\\
Moshika-RL & $K_L=4$; queue depth 3; persistent frame queue; prefix validation and suffix repair.\\
BayLing-Duplex & Five-frame future-block horizon; exact-prefix commit in the FP32 flattened token stream.\\
PersonaPlex & $K_L=4$; queue depth 3; immediate playout with repaired model state.\\
\bottomrule
\end{tabular}
\end{table}

\FloatBarrier
\section{HaluEval FBR Protocol and Additional Analysis}
\label{app:closed-loop}
The HaluEval evaluation continues generation after conditioning on admitted
references and scores the generated speech.

Speech inference and retrieval share one A100. The native reference encoder
runs as the same external A100 service for Dense and Joint. Its latency is
included in the admission deadline. Thus, the experiment matches the complete service allocation across
methods. The conditioner consumes
reference embeddings in order, with its consumption offset restored on
rollback. Reference completion alone is not an injection: the reference must
be encoded, admitted, and consumed by subsequent Temporal forwards.
Retrieval, reference encoding, and injection use the available frame budget,
with one extra frame allowed at the model's retrieval trigger.
\paragraph{Scoring.}
Whisper large-v3-turbo transcribes the actual generated audio. A local
Qwen2.5-14B-Instruct judge uses the retained dataset-specific answer rubric,
greedy decoding, and identical prompts without method labels or timing
information. The score measures the correctness of the required answer, not
factual correctness of every additional sentence. Dedicated evidence scoring
separately checks consumed references. Normalized answer-alias matching is
retained as a lexical diagnostic and increases from 73 to 175 on HaluEval.

The 500 HaluEval pairs contain 81 both-correct, 124 Joint-only-correct,
5 Dense-only-correct, and 290 both-incorrect outcomes.

\end{document}